\pdfoutput=1

\documentclass[11pt]{article}

\usepackage[preprint]{acl}

\usepackage{times}
\usepackage{latexsym}
\usepackage{graphicx} 
\usepackage{multirow}
\usepackage[T1]{fontenc}

\usepackage[utf8]{inputenc}

\usepackage{microtype}

\usepackage{inconsolata}

\usepackage{textcomp}
\usepackage{xspace}
\usepackage{tabularx}
\usepackage{url}
\usepackage{footmisc}
\usepackage{subcaption}
\usepackage{amsmath}

\newcommand{\method}[0]{LSTPrompt\xspace}

\newcommand{\moduleA}[0]{TimeDecomp\xspace}
\newcommand{\moduleB}[0]{TimeBreath\xspace}
\usepackage{todonotes}
\usepackage{xcolor}

\usepackage{listings}
\usepackage{xcolor}
 
\definecolor{codegreen}{rgb}{0,0.6,0}
\definecolor{codegray}{rgb}{0.5,0.5,0.5}
\definecolor{codepurple}{rgb}{0.58,0,0.82}
\definecolor{backcolour}{rgb}{0.95,0.95,0.95}

\lstdefinestyle{mystyle}{
    backgroundcolor=\color{backcolour},   
    commentstyle=\color{codegreen},
    keywordstyle=\color{magenta},
    numberstyle=\tiny\color{codegray},
    stringstyle=\color{codepurple},
    basicstyle=\ttfamily\footnotesize,
    breakatwhitespace=false,         
    breaklines=true,                 
    captionpos=f,                    
    keepspaces=true,                 
    numbers=left,                    
    numbersep=5pt,                  
    showspaces=false,                
    showstringspaces=false,
    showtabs=false,                  
    tabsize=2
}
 
\title{LSTPrompt: Large Language Models as Zero-Shot Time Series Forecasters by Long-Short-Term Prompting}

\author{Haoxin Liu\thanks{Equal Contribution}\textsuperscript{,\textdagger}, Zhiyuan Zhao\footnotemark[1]\textsuperscript{,\textdagger}, Jindong Wang\textsuperscript{\S},  Harshavardhan Kamarthi\textsuperscript{\textdagger}, B. Aditya Prakash\textsuperscript{\textdagger} \\ \textsuperscript{\textdagger}Georgia Institute of Technology, \textsuperscript{\S}Microsoft Research Asia\\ \textsuperscript{\textdagger}\text{\{hliu763, leozhao1997, hkamarthi3, badityap\}@gatech.edu}, \textsuperscript{\S}\text{jindong.wang@microsoft.com}}

\begin{document}
\maketitle
\begin{abstract}
Time-series forecasting (TSF) finds broad applications in real-world scenarios. 
Prompting off-the-shelf Large Language Models (LLMs) demonstrates strong zero-shot TSF capabilities while preserving computational efficiency. However, existing prompting methods oversimplify TSF as language next-token predictions, overlooking its dynamic nature and lack of integration with state-of-the-art prompt strategies such as \textit{Chain-of-Thought}. Thus, we propose \method, a novel approach for prompting LLMs in zero-shot TSF tasks. \method decomposes TSF into short-term and long-term forecasting sub-tasks, tailoring prompts to each. \method guides LLMs to regularly reassess forecasting mechanisms to enhance adaptability. Extensive evaluations demonstrate consistently better performance of \method than existing prompting methods, and competitive results compared to foundation TSF models\footnote{\url{https://github.com/AdityaLab/lstprompt}}.


\end{abstract}

\section{Introduction}
\label{sec:intro}








Time-series (TS) data are ubiquitous across various domains, including public health~\cite{epideep}, 
finance~\cite{d2}, and energy~\cite{d3}. Time-series forecasting (TSF), a crucial task in TS data analysis, aims to predict future events or trends based on historical data. Recent advancements in large Pre-Trained Models (PTMs), a.k.a. foundation models, and Large Language Models (LLMs) have demonstrated their effectiveness for TSF tasks. This is achieved either by training TS foundation models from scratch~\cite{ptm1, lptm, timegpt, timefm} or adapting LLMs to TS data as natural language modalities~\cite{timellm, llm4ts, promptcast, llmtime}. These methods leverage powerful generalization capabilities of PTMs or LLMs, proving effectiveness in zero-shot TSF tasks with promising applications without the need for domain-specific training data.

\begin{figure}[t]
  \centering
  \vspace{-0.0in}
  \includegraphics[width=0.49\textwidth]{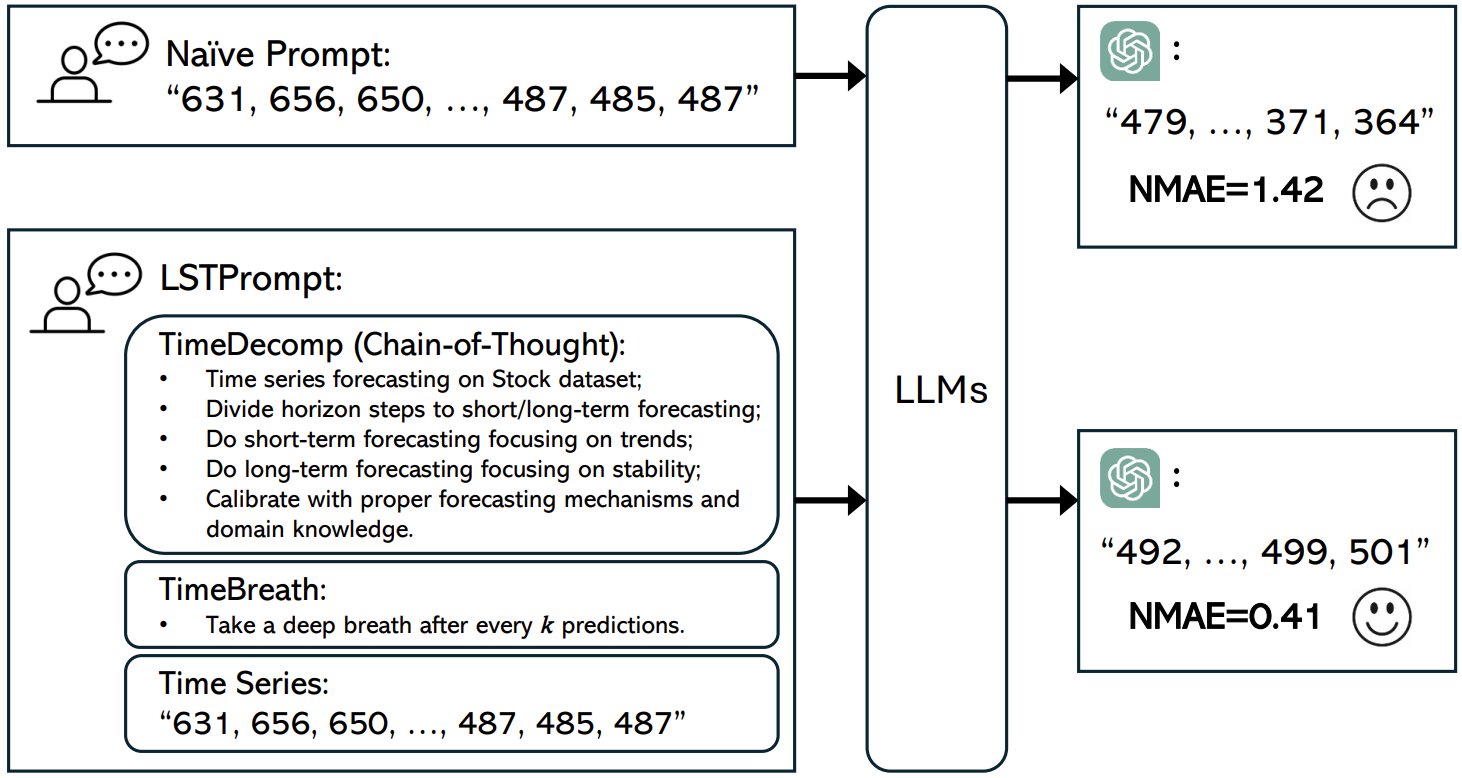} 
  \vspace{-0.3in}
  \caption{Comparison between naive prompt~\cite{llmtime} and \method. 
  }
  \label{fig:demo}
  \vspace{-0.1in}
\end{figure}


Designing proper prompting techniques for zero-shot TSF tasks offers notable advantages, which avoids training models from scratch or fine-tuning LLMs for computational efficiency while maintaining forecasting accuracy. Existing approaches~\cite{promptcast, llmtime} prompt LLMs for zero-shot TSF tasks by aligning TS data with natural language sequences and prompting LLMs to perform TSF as sequence completion tasks. However, these methods overlook the dynamic nature of TS data and the intricate forecasting mechanisms inherent in TSF tasks, such as modeling temporal dependencies, which cannot be adequately modeled by simple sequence completion tasks.


To address the limitation, we introduce \method, a novel prompt strategy of LLMs for TSF tasks by providing specific TSF-oriented guidelines. Our contributions are summarized as follows:
\begin{itemize}
    \item We propose \underline{L}ong-\underline{S}hort-\underline{T}erm Prompt (\method), which decomposes TSF into short-term and long-term forecasting subtasks. Each subtask guides LLMs with distinct forecasting rules and mechanisms, forming a \textit{Chain-of-Thought} reasoning path for predictions. 
    \item We introduce \moduleB to \method, an innovative component that encourages LLMs to regularly revisit forecasting mechanisms, enabling leveraging different forecasting mechanisms for different time periods.
    \item We evaluate \method on multiple benchmark and concurrent datasets, demonstrating its effectiveness for zero-shot TSF tasks. We show its generalization ability to outperform non-zero-shot methods in specific scenarios.
\end{itemize}
We provide additional related works in the Appendix \ref{sec:app_related} with distinguishing the differences of popular zero-shot TSF methods in Table \ref{tbl:sum}.

\section{Methodology}
\label{sec:method}

\subsection{Problem Formulation and Motivation}
Zero-shot TSF aims to predict future TS $\{y_i\}_{i=t}^{t+H}$ with a horizon window size $H$ based on a reference TS $\{y_i\}_{i=t-L}^t$ with lookback window size $L$, without prior exposure or training on the target series. Solving zero-shot TSF tasks with LLMs requires aligning TS data with natural language modalities to leverage remarkable generalization abilities and generate predictions based on the provided context.

One approach to align TS data with LLMs is to present TS data as text. Existing zero-shot TSF prompt strategies~\cite{promptcast, llmtime} represent TS data as strings of numerical digits and treat TSF tasks as text-based next-token predictions. 
However, these strategies overlook the need for sophisticated forecasting mechanisms inherent in dynamic TS data. Without explicit instructions, existing strategies may yield inaccurate predictions with high uncertainty.



To address this, we propose \method, tailored for zero-shot TSF tasks through prompting LLMs informatively. \method comprises two components: (1) \textbf{\moduleA}, decomposing TSF tasks into subtasks for systematic reasoning, and (2) \textbf{\moduleB}, facilitating periodic breaks to adapt forecasting strategies within the horizon window. We detail each module in the subsequent sections.



\subsection{\moduleA}



Rather than directly prompting complex questions to LLMs, recent studies advocate decomposing inquiries into simpler, sequential steps~\cite{cot, stepbystep}. This approach aids LLMs in constructing a coherent reasoning path. However, applying such chain-of-thought or step-by-step strategies to TSF tasks remains unexplored.

To address this, we introduce \moduleA, which breaks down TSF tasks into short-term and long-term forecasting subtasks. This is motivated by different forecasting mechanisms for short/long-term forecasting.  
Particularly, \moduleA prompts LLMs to partition horizon time steps into short-term and long-term accordingly. Then, it guides LLMs through each subtask, directing them to focus on specific aspects: short-term forecasting emphasizes trend changes and dynamic patterns, while long-term forecasting highlights statistical properties and periodic patterns. \moduleA's chain-of-thought process follows step-by-step cues: it prompts tasks with specific datasets, decomposes tasks into short-term and long-term sub-tasks, and guides LLMs to incorporate appropriate forecasting mechanisms and domain knowledge.



\subsection{\moduleB}

In addition to chain-of-thought prompting, recent studies emphasize the importance of incentivizing LLMs to follow step-by-step reasoning, especially when having numerous subtasks~\cite{breath_ape, breath_opro}. To facilitate this, \citeauthor{breath_opro} propose a strategy that introduces "Take a deep breath" before initiating step-by-step tasks.


TSF tasks involve varying reasoning across different time steps and overly lengthy forecasting horizons can overwhelm LLMs' reasoning abilities. Inspired by the "deep breath" design, we introduce \moduleB, which prompts LLMs to take "rhythmic breaths" during sequential reasoning for TSF. In the TSF task with $H$ time steps horizon, \moduleB guides LLMs to rhythmically breathe
every $k$ steps, where $k$ is a hyperparameter determining the breath frequency. The intuition of \moduleB is to encourage LLMs to reassess forecasting mechanisms regularly, particularly for distant time steps that may require different reasonings. By taking breaks, \moduleB helps LLMs avoid prior irrelevant inferences and fosters adaptive forecasting mechanisms to current forecasts.


In practice, the choice of $k$ significantly impacts LLMs' performance in zero-shot TSF tasks, as demonstrated in the sensitivity analysis provided in Appendix \ref{sec:app_exp}. A straightforward approach is to align the frequency of breaks with the upper time scale. For example, setting $k=5$ prompts weekly breaks for daily stock forecasting, while $k=4$ encourages monthly breaks for weekly Influenza forecasting.

\begin{table*}[t]
\centering
\vspace{-0.1in}
\resizebox{\linewidth}{!}{
\begin{tabular}{cccc|cccc|c|cc}
\hline
\multirow{6}{*}{Darts} & \multirow{2}{*}{Dataset} & \multirow{2}{*}{Frequency} & \multirow{2}{*}{Horizon} & \multicolumn{4}{c|}{Supervised} & \multicolumn{1}{c|}{\begin{tabular}[c]{@{}c@{}} Zero-Shot\\ (PTMs)\end{tabular}} & \multicolumn{2}{c}{\begin{tabular}[c]{@{}c@{}} Zero-Shot\\ (Prompt)\end{tabular}} \\ \cline{5-11}
& & & & SP & ARIMA & TCN & N-BEATS & TimesFM & LLMTime & \method \\  \cline{2-11}
& AirPassengers & Month & 29 & 34.67 & \underline{\textit{24.03}} & 54.96 & 97.89 & 14.75 & 48.96 & \textbf{13.02} \\
& MilkProduction & Month & 34 & \underline{\textit{30.33}} & 37.19 & 70.86 & 33.64 & 22.46 & 63.15 & \textbf{7.71} \\
& BeerProduction & Season & 43 & 102.05 & 17.13 & 30.90 & \underline{\textit{10.39}} & \textbf{10.25} & 20.85 & \underline{13.29} \\
& Sunspots & Day & 141 & 53.74 & \underline{\textit{43.56}} & 51.82 & 73.15 & 50.88 & 59.91 & \textbf{46.84} \\ 
\hline
\multirow{3}{*}{Monash} & & & & DeepAR & N-BEATS & WaveNet & Transformer & TimesFM & LLMTime & \method  \\ \cline{2-11}
& RiverFlow & Day & 30 & 23.51 & 27.92 & \underline{\textit{22.17}} & \multicolumn{1}{c|}{28.06} & 24.53 & 28.63 & \textbf{24.17} \\
& US Births & Day & 30 & 424.9 & \underline{\textit{422.0}} & 504.4 & \multicolumn{1}{c|}{452.9} & \textbf{408.5} & 459.43 & \underline{429.2} \\ 
\hline
\multirow{7}{*}{\begin{tabular}[c]{@{}c@{}}Informer\\ (ETT)\end{tabular}} & & & & Informer & Autoformer & FEDformer & PatchTST & TimesFM & LLMTime & \method \\ \cline{2-11}
& \multirow{2}{*}{ETTh1} & \multirow{2}{*}{Hour} & 96 & 0.76 & 0.55 & 0.58 & \underline{\textit{0.41}} & 0.37 & 0.42 & \textbf{0.32} \\
& & & 192 & 0.78 & 0.64 & 0.64 & \underline{\textit{0.49}} & 0.49 & 0.50 & \textbf{0.36} \\
& \multirow{2}{*}{ETTm1} & \multirow{2}{*}{Minute} & 96 & 0.71 & 0.54 & 0.41 & \underline{\textit{0.33}}  & 0.25 & 0.37  & \textbf{0.19} \\
& & & 192 & 0.68 & 0.46 & 0.49 & \underline{\textit{0.31}} & \textbf{0.24} & 0.71 & \underline{0.55} \\
& \multirow{2}{*}{ETTh2} & \multirow{2}{*}{Hour} & 96 & 1.94 & 0.65 & 0.67 & \underline{\textit{0.28}} & \textbf{0.28} & 0.33 &{\underline{0.31}} \\
& & & 192 & 2.02 & 0.82 & 0.82 & \underline{\textit{0.68}}  & 0.58 & 0.70 & \textbf{0.45} \\ \hline
\end{tabular}}
\vspace{-0.1in}
\caption{Performance comparison of supervised models and zero-shot methods on benchmark datasets: (1) \method achieves mostly the best and several second-best results among zero-shot forecasting methods. (2) \method outperforms the best supervised models on 6 out of 12 datasets.
We bold the best zero-shot results and \method with the second-best results is underlined. We italicize/underline the best supervised results.
}
\label{tbl:bench}
\end{table*}

\begin{table*}[t]
\centering
\vspace{-0.1in}
\resizebox{0.78\linewidth}{!}{
\begin{tabular}{ccc|cccc|c|cc}
\hline
\multirow{2}{*}{Dataset} & \multirow{2}{*}{Frequency} & \multirow{2}{*}{Horizon} & \multicolumn{4}{c|}{Supervised} & \multicolumn{1}{c|}{\begin{tabular}[c]{@{}c@{}} Zero-Shot\\ (PTMs)\end{tabular}} & \multicolumn{2}{c}{\begin{tabular}[c]{@{}c@{}}Zero-Shot\\ (Prompt)\end{tabular}} \\ \cline{4-10}
& & & Informer & AutoFormer & FedFormer & PatchTST & LPTM & LLMTime & \method \\ \hline
\multirow{4}{*}{ILI} & \multirow{4}{*}{Week} & 4 & 1.64 & 1.17 & 2.31 & \underline{\textit{0.51}} & 1.54 & 0.61 & \textbf{0.42} \\
& & 12 & 2.25 & 2.10 & 1.97 & \underline{\textit{0.52}} & 0.83 & 0.81 & \textbf{0.67} \\
& & 20 & 2.01 & 1.43 & 1.67 & \underline{\textit{1.39}} & 1.70 & 4.68 & \textbf{1.73} \\
& & 24 & 4.29 & 1.86 & \underline{\textit{1.30}} & 2.15 & 2.18 & 4.81 & \textbf{2.08} \\ \hline
\multirow{4}{*}{Stock} & \multirow{4}{*}{Day} & 24 & 5.07 & 9.94 & 8.73 & \underline{\textit{4.52}} & 0.73 & 0.51 & \textbf{0.32} \\
& & 48 & 8.03 & 9.22 & 9.56 & \underline{\textit{4.11}} & 0.80 & 0.42 & \textbf{0.19} \\
& & 96 & \underline{\textit{3.11}} & 9.61 & 9.43 & 4.36 & 0.87 & 1.42 & \textbf{0.41} \\
& & 120 & \underline{\textit{4.07}} & 10.92 & 10.59 & 4.65 & 1.28 & 2.61 & \textbf{0.52} \\ \hline
\multirow{4}{*}{Weather} & \multirow{4}{*}{Day} & 24 & 1.59 & \underline{\textit{1.54}} & 1.77 & 1.77 & 0.79 & \textbf{0.31} & \textbf{0.31} \\
& & 48 & 1.62 & 1.63 & 1.84 & \underline{\textit{1.25}} & 1.06 & 0.66 & \textbf{0.53} \\
& & 96 & 1.43 & 1.50 & 2.34 & \underline{\textit{1.16}} & 1.08 & 0.84 & \textbf{0.62} \\
& & 120 & 1.45 & 1.64 & 1.95 & \underline{\textit{1.40}} & 1.18 & 0.83 & \textbf{0.69} \\ \hline
\end{tabular}}
\vspace{-0.1in}
\caption{Performance comparison of supervised models and zero-shot methods on concurrent datasets: (1) \method consistently outperforms zero-shot baselines on all evaluations. (2) \method outperforms best supervised models on 9 of 12 evaluations. We bold the best zero-shot method and italicize/underline the best supervised results.
}
\vspace{-0.1in}
\label{tbl:zero}
\end{table*}

\subsection{\method}
We introduce \method, which integrates \moduleA and \moduleB to create the comprehensive prompt strategy. The prompt is straightforward: \method first guides LLMs through the chain-of-thought steps outlined by \moduleA, then instructs them to take regular breaks using \moduleB. A \method demo is shown by Figure \ref{fig:demo}. We provide a detailed prompting example in Appendix \ref{sec:app_prompt}. \method is designed for any TS datasets for zero-shot TSF tasks. It can be easily tailored to different scenarios by adjusting a single hyperparameter, $k$, as previously discussed.
\section{Experiments}
\label{sec:exp}

\subsection{Benchmark Evaluation}
To benchmark the performance of \method, we use three common TSF benchmarks: Darts~\cite{darts}, Monash~\cite{monash}, and Informer datasets~\cite{informer}. 
While these datasets can potentially be used for training LLMs, evaluating \method on these datasets allows fair comparisons within aligned settings, which strictly follows the established setup for zero-shot TSF tasks~\cite{llmtime} and are detailed in Appendix \ref{sec:app_exp}. We use the SOTA prompting method LLMTime~\cite{llmtime} and a recent PTM  TimesFM~\cite{timefm} as zero-shot baselines. 
The results are shown in Table \ref{tbl:bench}. We showcase visualized results in Appendix \ref{sec:app_exp}.

The results highlight two main benefits of \method: First, \method achieves the best performance on 8 out of 12 benchmark datasets and the second-best performance on the remaining 4 among zero-shot methods. Notably, \method always outperforms the SOTA prompt method LLMTime, while may slightly lag behind TimesFM, which is expected since TimesFM is a TSF-specific PTMs.
Second, \method can outperform best supervised results under certain scenarios. For instance, \method achieves a 74.6\% lower MAE compared to the best supervised result on the MilkProduction dataset. This improvement relies on the strong generalization ability of LLMs, which helps mitigate overfitting for supervised models.

\subsection{Concurrent Dataset Evaluation}
To evaluate the true zero-shot ability of \method, we conduct experiments over three concurrent datasets from different domains: influenza-like illness (ILI), Stock, and Weather (Detailed in Appendix \ref{sec:app_exp}). These datasets ensure that the test data are after June 2023,
while most LLMs are trained only up to 2022~\cite{gpt4}. Employing these datasets ensures the zero-shot property, even for GPT4. The experiment setup follows Benchmark Evaluations. 
We omit PromptCast~\cite{promptcast}, exclude TimesFM, and include another foundation time-series model, LPTM~\cite{lptm}, for zero-shot baselines with explanations in Appendix \ref{sec:app_exp}.
We include supervised TSF models, including Informer~\cite{informer}, Autoformer~\cite{autoformer}, FEDformer~\cite{fedformer}, and PatchTST~\cite{patchtst}, to show performance disparities between zero-shot methods and supervised models on TSF tasks. The results are shown in Table \ref{tbl:zero}.

The results demonstrate that \method consistently outperforms zero-shot baselines on all evaluations. Notably, \method consistently outperforms best supervised results on Stock and Weather datasets. This is attributed to heavy distribution drifts on these datasets, which largely degrade the supervised models' performances.
In contrast, benefiting from strong generalization abilities of LLMs and zero-shot properties,
zero-shot methods mitigate the impacts of distribution drifts and achieve better performance than supervised models.


\subsection{Ablation Study}

\begin{figure}[htbp]
  \centering
  \vspace{-0.1in}
  \includegraphics[width=0.45\textwidth]{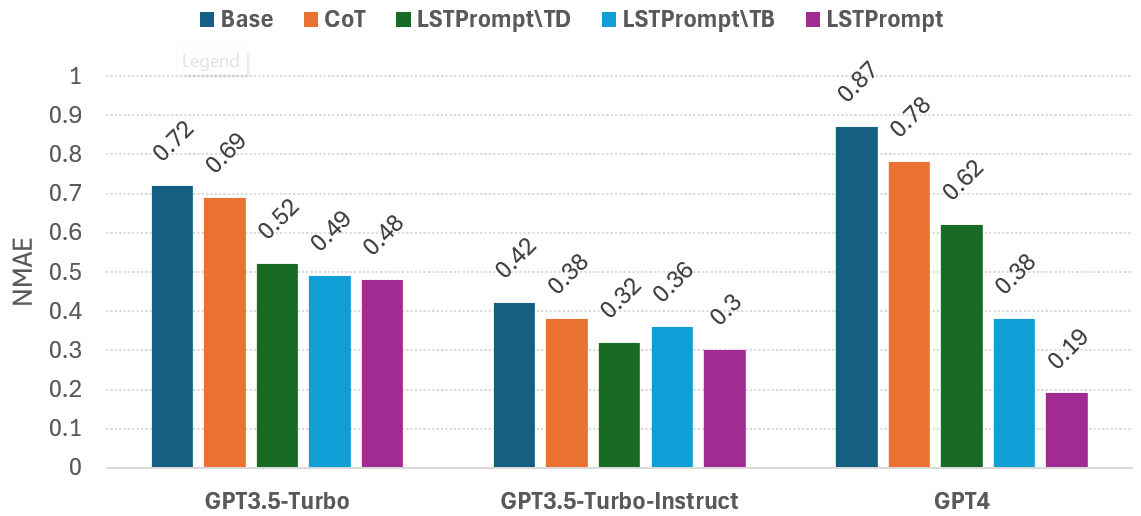} 
  \vspace{-0.1in}
  \caption{Ablation Study: (1) Enhanced reasoning abilities enable \method to perform best on GPT4. (2) Both \moduleA and \moduleB effectively enhance the forecasting accuracy of \method.}
  \label{fig:ablation}
  \vspace{-0.1in}
\end{figure}
\par\noindent To understand the significance of various components of \method, we conduct two ablation studies: (1) Analyzing the impact of employing different LLMs; (2) Analyzing the effects of \moduleA and \moduleB. We conduct experiments with combinations of different LLMs and various ablated versions of \method on the Stock dataset, with results visualized in Figure \ref{fig:ablation}.
\par\noindent\textbf{Prompting Different LLMs.}  In prior experiments, we presented forecasting results based on the most suitable LLMs (e.g., GPT3.5-Turbo-Instruct for LLMTime 
and GPT4 for \method). However, performance differences can arise among zero-shot TSF methods, including \method, when evaluated across different LLMs. Thus,we investigate and interpret the potential impacts of utilizing GPT3.5-Turbo, GPT3.5-Turbo-Instruct, and GPT4 with \method. The results indicate \method coupled with GPT4.0 outperforms instances with GPT3.5-Turbo and GPT3.5 Turbo-Instruct. 
This finding aligns with expectation, as \method prompts LLMs to follow the reasoning path through distinct short-term and long-term forecasting subtasks, each requiring different reasoning mechanisms, while GPT4 is known for its reasoning abilities compared to the remaining two.

\par\noindent\textbf{Module Effectiveness.} To understand the significance of \moduleA and \moduleB, we analyze performance discrepancies over three ablated versions of \method: (1) Base, using standard prompts; (2) LSTPrompt\textbackslash TD, excluding \moduleA from \method; (3) LSTPrompt\textbackslash TB, excluding \moduleB from \method. We include the state-of-the-art Chain-of-Thought method~\cite{breath_opro} (referred to as `CoT') to highlight performance differences with the SOTA prompt strategy for general tasks.

The results demonstrate the effectiveness of both \moduleA and \moduleB. Incorporating \moduleA and \moduleB reduces the average NMAE by 26.8\% and 34.1\%, respectively, compared to Base prompts. Employing both modules enhances average performance by 46.7\% than Base prompts. Moreover, the sole utilization of either \moduleA or \moduleB demonstrates certain advantages in forecasting accuracy over the best CoT method, highlighting the necessity of designing tailored prompts for TSF tasks.

\section{Conclusion}
\label{sec:conclusion}

In this paper, we introduce \method, a novel prompt paradigm for zero-shot TSF tasks through prompting LLMs. \method enables LLMs to achieve accurate zero-shot TSF tasks through two innovative modules: \moduleA, which decomposes zero-shot TSF tasks into a series of chain-of-thought subtasks, and \moduleB, which encourages LLMs to periodically reassess forecasting mechanisms. Extensive experiments validate the effectiveness of \method, which consistently outperforms the SOTA prompt method and shows generally better performance than SOTA PTMs.

\section{Acknowledgement}

This paper was supported in part by the NSF (Expeditions CCF-1918770, CAREER IIS-2028586, Medium IIS-1955883, Medium IIS-2106961, PIPP CCF-2200269), CDC MInD program, Meta faculty gift, and funds/computing resources from Georgia Tech and GTRI.

\bibliography{custom}

\clearpage

\appendix

\section{Additional Related Works}
\label{sec:app_related}

\par\noindent\textbf{Time-Series Forecasting.} Traditional time series methods approach forecasting from a statistical standpoint, treating it as standard regression problems with time-varying parameters~\cite{regression, gaussian, arima}. Recent advancements in deep learning have led to significant breakthroughs in this field, exemplified by deep models like LSTNet and N-BEATS~\cite{lstnet, nbeats}. Many state-of-the-art deep learning methods, such as Informer, Autoformer, PatchTST, and CAMul~\cite{informer, autoformer, patchtst, camul}, build upon the success of self-attention mechanisms, popularized by transformer-based architectures~\cite{transformer}. These transformer-based models excel at capturing long-range dependencies, surpassing the capabilities of traditional Recurrent Neural Network (RNN) models, owing to their effective use of self-attention mechanisms.
\vspace{0.05in}

\par\noindent\textbf{Large Language Models.} The augmentation of language model parameters and training data size has been shown to enhance generalization ability~\cite{gpt}. Consequently, researchers have developed Large Language Models (LLMs) like GPT~\cite{gpt, gpt4} and Llama~\cite{llama2}. These models excel at identifying patterns in prompts and extrapolating them through next-token prediction, achieving remarkable success in few-shot or zero-shot generalization and in-context learning. Beyond natural language tasks, LLMs exhibit effectiveness in transfer learning across diverse modalities, including images~\cite{llm4image}, audio~\cite{llm4audio}, tabular data~\cite{llm4table}, and time-series data~\cite{llm4tsd}. These accomplishments underscore the importance of aligning modalities appropriately to enable LLMs to comprehend tokenized patterns across different domains beyond traditional language processing tasks.

\vspace{0.05in}

\par\noindent\textbf{Large Models for Time-Series Forecasting.} In addition to the success of large models in language tasks, researchers in the field of time-series forecasting (TSF) have pursued the development of large models from two main perspectives: First, they train Pre-Trained Time-Series Models from scratch~\cite{timegpt, timefm, lptm, ptm1}, utilizing extensive time-series datasets and tailoring them specifically for TSF tasks. Alternatively, researchers harness the generalization capabilities of Large Language Models (LLMs) by aligning time-series data with language modalities through techniques such as reprogramming~\cite{timellm, llm4ts, llm4tsd} or prompting~\cite{llmtime, promptcast}.\begin{table}[h]
\centering
\resizebox{\linewidth}{!}{
\begin{tabular}{ccccc}
\hline
Method & Type &  Cost & \begin{tabular}[c]{@{}c@{}}Use CoT or\\ Guidelines\end{tabular} & \begin{tabular}[c]{@{}c@{}}Evaluated \\ on GPT4\end{tabular} \\ \hline
TimesFM    & PTMs & High & N/A  & N/A \\
LPTM       & PTMs & High & N/A  & N/A \\ \hline
PromptCast & Prompt & Low & No & No \\
LLMTime    & Prompt & Low & No & Partial \\ 
\method       & Prompt & Low & Yes & Yes   \\ \hline                           
\end{tabular}}
\caption{Summary of similarities and differences of related works on zero-shot TSF tasks.}
\label{tbl:sum}
\end{table}
To better understand the similarities and differences between all zero-shot methods mentioned in this work, we list the property comparisons of all zero-shot methods in Table \ref{tbl:sum}. 

\vspace{0.05in}
\section{Prompt Details}
\label{sec:app_prompt}

Below, we introduce a template prompt for \method, designed to be adaptable to various time-series datasets for zero-shot time-series forecasting tasks. The template is outlined as follows:

\begin{lstlisting}[caption=\method prompts, label=ls:gno.ner.prompts]
f"Please continue the following input sequence by addressing the task of forecasting {dataname}. You should break down the task into short-term and long-term predictions, following a three-step plan. First, adaptively and reasonably identify the ranges for short-term and long-term predictions. Then, design distinct and correct forecasting mechanisms for both short-term and long-term prediction tasks. For short-term predictions, focus on trends and the last few steps of the input sequence. For long-term predictions, emphasize cyclical patterns and statistical properties of the entire input sequence. You may further optimize the forecasting mechanisms based on your observations and domain knowledge. Finally, correctly implement the forecasting mechanisms, completing predictions one-time step at a time.
Remember to take a deep breath after every {breath_steps} time steps of prediction. The input sequence is as follows:\n"
\end{lstlisting}
\section{Additional Experiment Details}
\label{sec:app_exp}

\par\noindent\textbf{Experiment Setup.} Following the established setups in LLMTime and with the consideration of evaluating costs, we limit our focus to univariate time series forecasting tasks. However, \method can readily extend to the multivariate forecasting domain by employing multiple univariate forecasting techniques~\cite{llmtime,M2}. We strictly followed LLMTime's data-splitting method for benchmark datasets, where the test set comprises the last 20\% of each time series.

In addition to well-known benchmark datasets such as Darts, Monash, and ETT, our zero-shot evaluations encompass three concurrent datasets: ILI, Stock, and Weather. This selection ensures that the test data have never been exposed to LLMs training. All these datasets are publicly accessible. We use data after June 2023 for testing, thereby guaranteeing that GPT-3.5 and GPT-4 models have not been trained on these sets. Further details on these datasets are provided below:
\begin{itemize}
    \item \textbf{ILI\footnote{\url{https://gis.cdc.gov/grasp/fluview/fluportaldashboard.html}}:} The ILI dataset provides the reported influenza-like illness patients with age divisions. The dataset covers from 2002 to 2023. The forecasting target is the weekly number of ILI patients. 
    \item \textbf{Stock\footnote{\url{https://www.kaggle.com/datasets/jillanisofttech/google-10-years-stockprice-dataset}}:} The Stock dataset provides daily historical data of Alphabet Inc. (GOOG). The Stock dataset set has 7 columns, including the stock's opening price, closing price, highest price of the day, etc. The dataset covers from 2013 to 2024 (Jan). The forecasting target is the daily opening price. 
    \item \textbf{Weather\footnote{\url{https://www.kaggle.com/datasets/curiel/chicago-weather-database}}:} The Weather dataset provides historical weather record of Chicago. This dataset set has 10 columns, including date, temperature, precipitation, humidity, wind speed, and atmospheric pressure. The dataset covers from 2021 to 2023. The forecasting target is the daily average temperature.
\end{itemize}

\vspace{0.05in}

\par\noindent\textbf{Baseline.} In supervised baselines, we adopt various models depending on the benchmark's official evaluation criteria. Concurrent datasets utilize transformer-based supervised models, same as the ETT benchmark, known for their remarkable performance in TSF evaluations.

For zero-shot baselines, we categorize methods into pre-trained Time-Series Foundation Models (PTMs) and prompting methods. In benchmark evaluations, we utilize TimesFM~\cite{timefm} for PTMs, as it asserts not being trained on these datasets, while LPTM~\cite{lptm} does. Conversely, for concurrent dataset evaluations, we employ LPTM, as it is open-source compared to TimesFM. Reprogramming methods are omitted, such as TimeLLM~\cite{timellm} and LLM4TS~\cite{llm4ts}, due to their inapplicability to our zero-shot setting. 

For prompting methods, we compare LLMTime~\cite{llmtime} with our proposed \method. Promptcast~\cite{promptcast} is omitted, as LLMTime consistently outperforms it, and \method demonstrates uniformly better performance across all evaluations than LLMTime.

\vspace{0.05in}

\par\noindent\textbf{Evaluation Metric.} Following the established setups, we evaluate the Mean Absolute Error (MAE) on Darts and Monash datasets between predictions and raw target sequences. For ETT, ILI, Stock, and Weather datasets, we evaluate the MAE based on the normalized predictions and target sequences according to the mean and variance of the training data. The formulation of MAE $ = \frac{1}{n} \sum_{t=1}^{n} |y_t - \hat{y}_t| $.

\vspace{0.05in}

\par
\begin{figure}[h]
  \centering
  \includegraphics[width=0.45\textwidth]{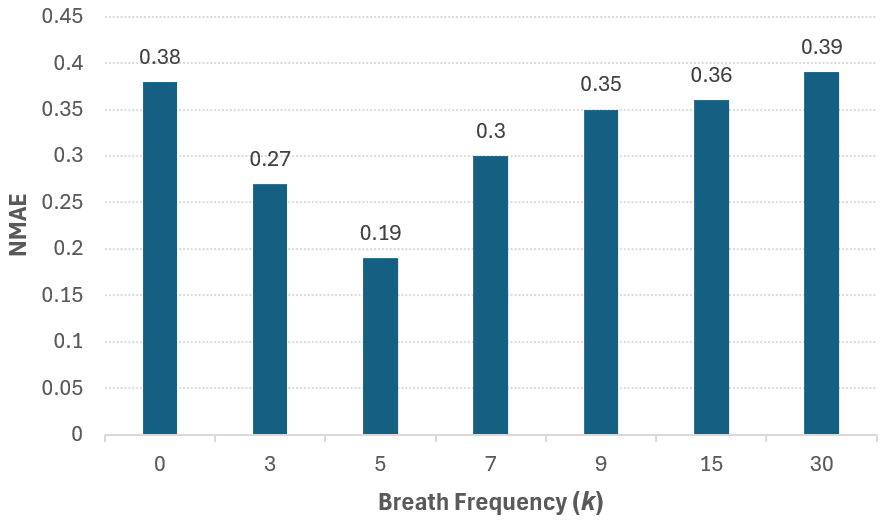} 
  \caption{Hyperparameter Sensitivity: The best breath frequency $k=5$ (weekly) aligns with the upper time scale of the Stock data (daily).}
  \label{fig:hyper}
\end{figure}

\noindent\textbf{Hyperparameter Sensitivity Study.} As previously mentioned in Section \ref{sec:method}, we conduct experiments using \method on the Stock dataset with varying values of breath frequency $k$. The results are shown in Fiugure \ref{fig:hyper}. Note that $k=0$ denotes \method without employing \moduleB.

The results suggest that setting $k=5$, enabling \method to breathe weekly in forecasting the stock prices, achieves the best performance compared to other breath frequencies. This optimal frequency aligns with the Stock dataset's structure, which includes daily stock prices for 5 weekdays. Intuitively, setting $k=5$ encourages \method to reassess its reasoning and forecasting strategy on a weekly basis, fitting well with the inherent weekly cycles in stock data. By appropriately adjusting the breath frequency in \moduleB, \method can dynamically infer patterns while effectively adapting to the data's periodic nature, leading to more accurate forecasts.

\begin{figure}[h]
    \centering
    \begin{subfigure}{0.45\textwidth}
        \centering
        \includegraphics[width=\linewidth]{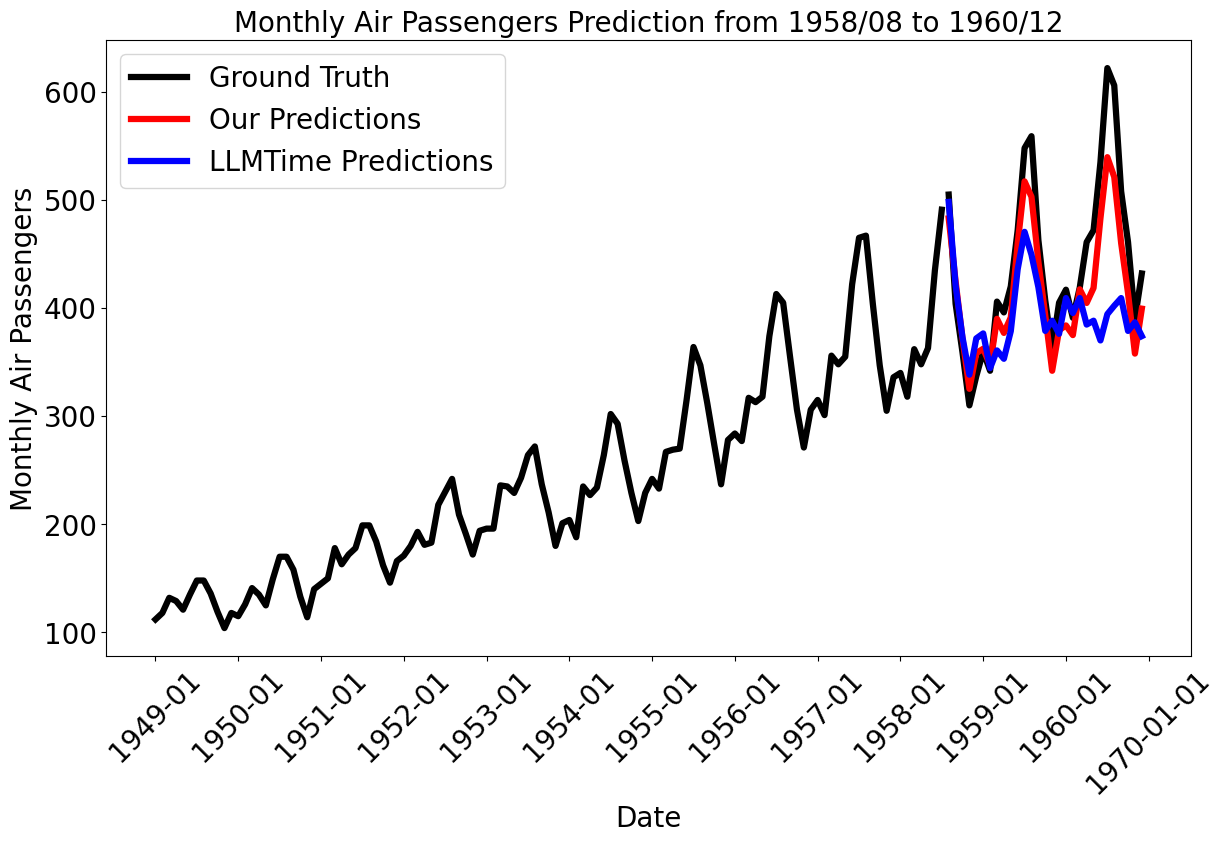}
    \end{subfigure}
    \begin{subfigure}{0.45\textwidth}
        \centering
        \includegraphics[width=\linewidth]{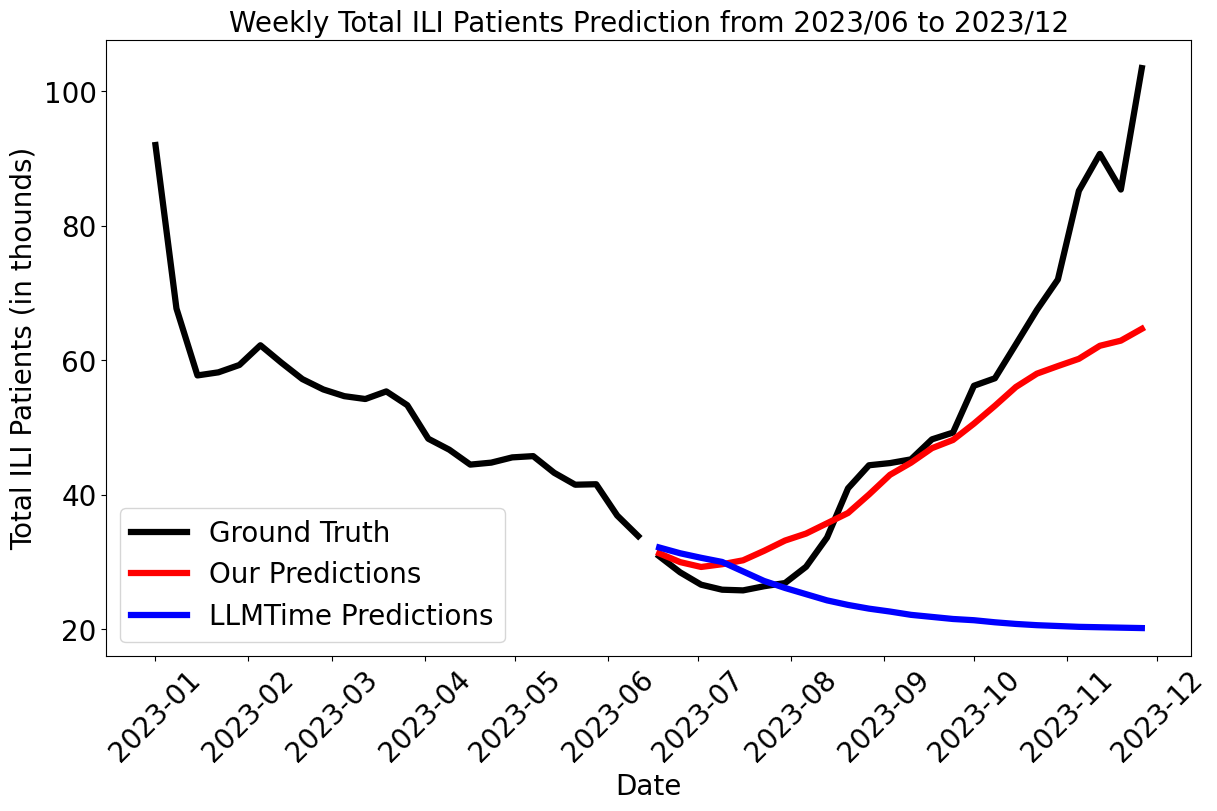}
    \end{subfigure}
    \vspace{-0.1in}
    \caption{Result visualizations on the AirPassengers (top) and ILI (bottom) datasets. \method exhibits better performance than LLMTime, demonstrating enhanced long-term prediction stability and improved ability to capture trend changes.}
    \label{fig:vis}
\end{figure}


\par\noindent\textbf{Result Visualization.} We present the result visualizations for the AirPassengers dataset in the Benchmark Evaluation and the ILI dataset in the Concurrent Dataset Evaluation. These visualizations are shown in Figure \ref{fig:vis}.

The visualizations demonstrate clear benefits from two perspectives: First, the predictions of \method exhibit greater long-term stability and accuracy compared to LLMTime, as evidenced by the AirPassengers predictions. Notably, \method effectively maintains the periodic properties inherent in the dataset. Second, \method demonstrates better capability in capturing accurate trend changes compared to LLMTime, as illustrated by the ILI predictions. In particular, \method accurately captures trends in increasing predictions where LLMTime fails to detect them.


\section{Limitation Discussion}
\label{sec:limit}

While \method has demonstrated effectiveness in zero-shot TSF tasks by employing simple prompts for LLMs, its limitations should be acknowledged from two perspectives. First, the interpretability of \method may be compromised. The evaluation of \method heavily relies on existing LLMs, the mechanisms and response behaviors of which are currently challenging to interpret. Consequently, \method may suffer from reduced interpretability due to our limited understanding of LLMs. Second, incorporating additional instructions in the prompts, such as the names and properties of time-series datasets, could potentially introduce information leaks that are exploited by the LLMs. We advocate for further research within the safe AI community to investigate the trustworthiness of LLMs, ensuring that \method can be deployed without concerns regarding information leakage issues.

\end{document}